\documentclass[fleqn, twocolumn,10pt]{wlscirep}
\usepackage[utf8]{inputenc}
\usepackage[T1]{fontenc}
\usepackage{comment}
\usepackage{subcaption}
\usepackage{multirow}
\usepackage{hyperref}
\hypersetup{
 breaklinks=true,
 colorlinks=true,
    linkcolor=blue,
    citecolor=blue,      
    urlcolor=blue
}
\usepackage{xcolor}
\usepackage{mdframed}
\usepackage{todonotes}
\graphicspath{{Figures/}}
\usepackage{lineno}

\newcommand{\COMMENT}[1]{}

\newcommand{\be}{\begin{equation}}
\newcommand{\ee}{\end{equation}}
\newcommand{\ben}{\begin{equation*}}
\newcommand{\een}{\end{equation*}}
\newcommand{\ba}{\begin{eqnarray}}
\newcommand{\ea}{\end{eqnarray}}
\newcommand{\ban}{\begin{eqnarray*}}
\newcommand{\ean}{\end{eqnarray*}}

\newcommand{\R}{\mathbb{R}}

\newcommand{\quark}{\setbox0\hbox{$x$}\hbox to\wd0{\hss$\cdot$\hss}}

\providecommand*{\DUrole}[2]{
  \ifcsname DUrole#1\endcsname
    \csname DUrole#1\endcsname{#2}
  \else
    \ifcsname docutilsrole#1\endcsname
      \csname docutilsrole#1\endcsname{#2}
    \else
      #2
    \fi
  \fi
}

\makeatletter
\g@addto@macro\@verbatim\footnotesize
\makeatother
\newif\ifshow
\showfalse
\ifshow
 
\else
 \excludecomment{orig}
\fi
\newif\ifshow
\showtrue
\ifshow
 
\else
 \excludecomment{mods}
\fi
\newif\ifshow
\showtrue
\ifshow
 
\else
 \excludecomment{edits}
\fi
\newif\ifshow
\showtrue
\ifshow
 
\else
 \excludecomment{notes}
\fi

\title{Efficient Learning of Accurate Surrogates for Simulations of Complex Systems} 

\author[1,*]{A.~Diaw}
\author[2,*]{M.~McKerns}
\author[2]{I.~Sagert}
\author[3]{L.~G.~Stanton}
\author[4]{M.~S.~Murillo}
\affil[1]{Fusion Energy Division, Oak Ridge National Laboratory, Oak Ridge, TN, USA}
 \affil[2]{Los Alamos National Laboratory, Los Alamos, NM, USA}
 \affil[3]{Department of Mathematics and Statistics, San Jos\'e State University, San Jos\'e, CA, USA}
\affil[4]{Department of Computational Mathematics, Science and Engineering, Michigan State University, East Lansing, MI, USA}

\affil[*]{corresponding: diawa@ornl.gov; mmckerns@lanl.gov}

\begin{abstract}
Machine learning methods are increasingly deployed to construct surrogate models for complex physical systems at a reduced computational cost. However, the predictive capability of these surrogates degrades in the presence of noisy, sparse, or dynamic data. We introduce an online learning method empowered by optimizer-driven sampling that has two advantages over current approaches: it ensures that all local extrema (including endpoints) of the model response surface are included in the training data and it employs a continuous validation and update process in which surrogates undergo retraining when their performance falls below a validity threshold. We find, using benchmark functions, that optimizer-directed sampling generally outperforms traditional sampling methods in terms of accuracy around local extrema even when the scoring metric is biased towards assessing overall accuracy. Finally, the application to dense nuclear matter demonstrates that highly accurate surrogates for a nuclear equation of state model can be reliably auto-generated from expensive calculations using few model evaluations.

\end{abstract}

\begin{document}


\flushbottom
\maketitle

\thispagestyle{empty}

Surrogate models play a pivotal role in science and engineering, offering practical solutions for modeling and optimizing systems where direct handling is computationally prohibitive. These models are particularly crucial in domains such as climate modeling, quantum information science, and automated instrumentation control, where they serve as efficient alternatives to costly simulations, enabling robust predictions \cite{Barros:2014, Wigley:2016, Scheinker:2015, Noack:2019}. A fundamental requirement for these surrogates is the assurance of accurately reflecting the system's behavior, especially in complex phenomena like phase transitions.

In materials science~\cite{Coveney}, macroscopic simulations heavily depend on surrogate models. These models integrate closure information derived from microphysical methods like molecular dynamics ~\cite{Hu2014, PhysRevX.8.021044} or Monte Carlo simulations~\cite{PhysRevLett.110.146405}. However, the challenge arises from the nature of the data used in these models, which is often high-dimensional, noisy, sparse, and may include time-dependent variables. The generation of adequate data to feed into these models usually demands an extensive number of microscopic calculations. This not only creates a significant computational burden but also poses potential barriers in the fields of materials discovery and design. Such limitations highlight the necessity for efficient and reliable surrogate models in predicting material properties with accuracy, as emphasized in recent studies~\cite{Schmidt:2019, Liu:2017}.

A prevalent technique in the development of surrogate models involves coupling the model's evaluation with an uncertainty metric. This metric assists in determining the need for additional fine-scale simulations. This methodology has been effectively utilized in various studies. For instance, Lubbers et al. \cite{Lubbers20} and Diaw et al. \cite{Diaw2020} have employed active learning strategies to develop surrogates that accurately represent fine-scale material responses. Similarly, Roehm et al. \cite{roehm2015distributed} adopted kriging, a statistical method, for constructing surrogates of stress fields. These surrogates were then integrated with a fine-scale code to address the macro-scale conservation laws in elastodynamics. In the realm of autonomous X-ray scattering experiments, Noack et al. ~\cite{Noack:2019} also implemented a kriging-based approach for surrogate creation. Their methodology included the use of a genetic algorithm to optimize the variances at each data point, subsequently drawing new samples from a distribution centered around these optimized maxima. However, a critical limitation observed in these studies is the lack of guaranteed validity of the surrogates for future data sets. This absence of assurance raises concerns about the long-term applicability and reliability of the surrogates, underscoring the need for continuous evaluation and adaptation of these models.

We propose an online learning methodology to efficiently construct surrogates that are \emph{asymptotically} valid for any future data
(see 'Surrogate Validity' in Methods); details of this claim are discussed in the modeling section. Our methodology has three key components: (1) a sampling strategy to generate new training and test data, (2) a learning strategy to generate candidate surrogates from the training data, and (3) a validation metric to evaluate candidate surrogates against the test data. Radial basis function (RBF) interpolation \cite{coulomb2003comparison,wu2012using,park1991universal} are used as the surrogate estimator's response surface. The numerical realization is done with \emph{mystic}, an open-source optimization, learning, and uncertainty quantification toolkit \cite{MHA_:2009, MSSFA:2011}. Our online approach is designed to include the first- and second-order critical points (including endpoints) in the search criteria, and thus, we conjecture that the training set needs only to include the critical points. The intended design should yield a negligible increase in retraining times.

\begin{figure}[!t]
\includegraphics[width=1\linewidth, height=0.2\textheight]{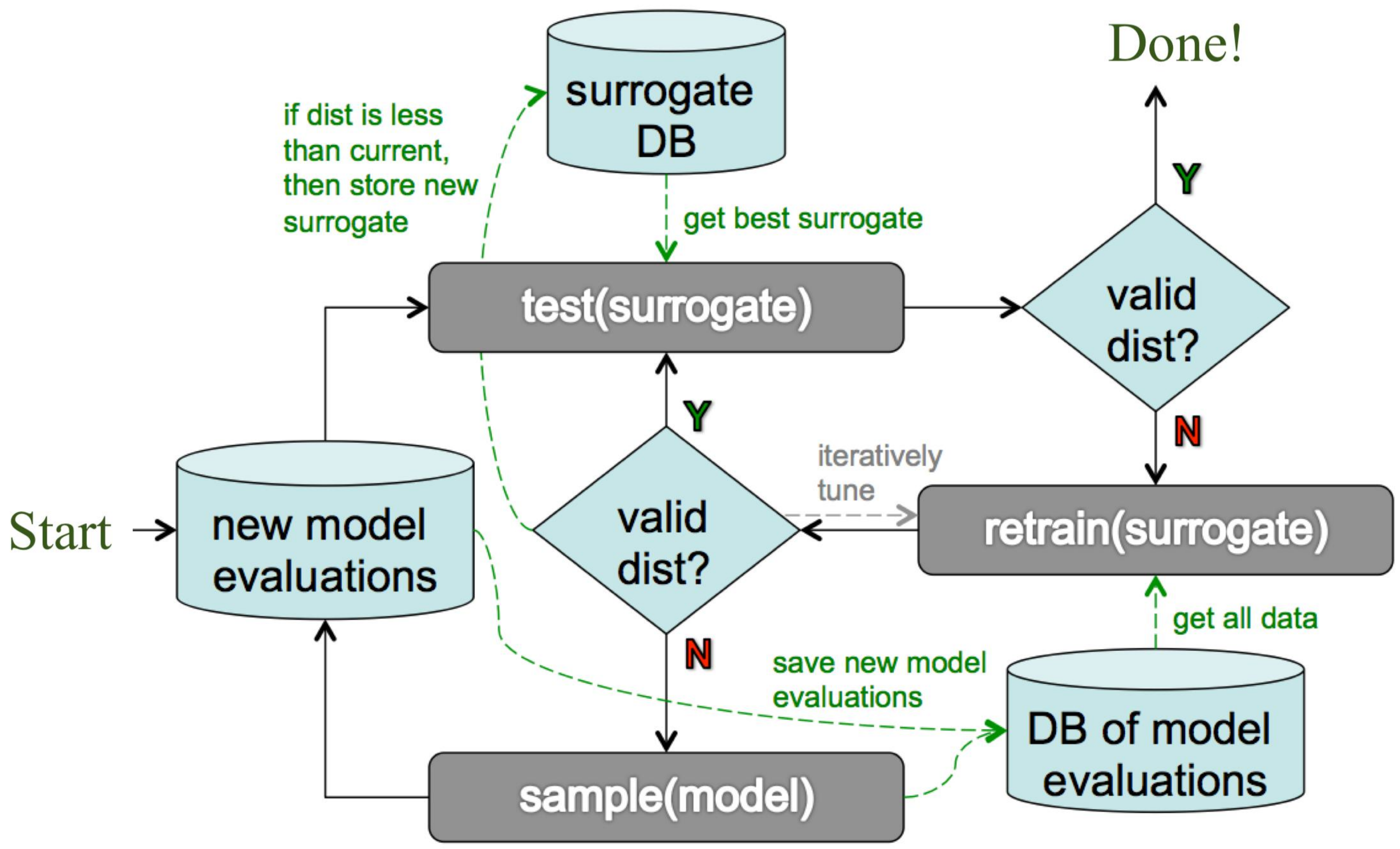}
    \caption{\textbf{Schematic for Automated Generation of Inexpensive Surrogates for Complex Systems}.
When new model evaluations occur, the corresponding surrogate 
is retrieved and evaluated for the same data. If the surrogate is determined to still be valid, the execution stops. Otherwise, the surrogate is updated by retraining against 
the DB of stored model evaluations, where the surrogate is validated with a fine-tuning of surrogate hyperparameters against a quality metric. If iterative retraining improves 
the surrogate, it is saved. Otherwise, new model evaluations are sampled to generate additional data. The process repeats until testing produces a valid surrogate.
}
    \label{fig:Valid-surrogate}
\end{figure}

The current work primarily focuses on how sampling strategy affects the efficiency when producing an asymptotically valid surrogate.
Here, we characterize validity via the evolution of the surrogate test score. %
Figure (\ref{fig:Valid-surrogate}) shows the procedure to create such a surrogate for an expensive model; 
it is iterative and includes explicit validation and update mechanisms.
To reduce computational complexity, we first link the model to a database (DB); thus, the model's input and output are automatically stored when it is evaluated. 
Later, the DB of model evaluations is used to train candidate surrogates. 
The corresponding surrogate is retrieved from the surrogate DB and tested for validity when the model is evaluated. 
If no stored surrogate exists, we skip testing and proceed directly to learning a candidate surrogate.

\section*{Results}

We first assess the performance of sampling strategies against benchmark functions, which is then extended to finding accurate surrogates for equation-of-state (EOS) calculations of dense nuclear matter. A second application of radial distribution functions in strongly coupled plasmas is given in the Supplement.

\subsection*{\label{sec:Applications} Numerical evaluation: Benchmark Functions}
Our case studies use benchmark functions commonly applied to test the performance of numerical optimization algorithms. We first examine how the sampling strategy affects the efficiency and effectiveness of finding an asymptotically valid surrogate. Then, we explore how the optimizer configuration impacts the efficiency of generating an initial valid surrogate.

\paragraph{\label{sec:TestStrategy} {\bf Sampling for Asymptotic Validity.}}
We first contrast the ability of optimizer-directed sampling with random sampling in finding an accurate surrogate with regard to all future data.
We use the workflow for asymptotic validity described in Section \ref{sec:SurrogateValidity} to learn a surrogate for the $d$-dimensional Rastrigin function \cite{Rastrigin:1974}:
\begin{equation}\label{eq:Rastrigin2}
    f({\bf x}) = 10 d + \sum^{d}_{i=1} \big[ x^{2}_{i} - 10 ~\cos(2\pi x_{i})\big],
\end{equation}
with $d = 2$, bounded by $x_{i} \in [0, 10]$, 
where the 2-D Rastrigin's function is essentially a spherical function with
a cosine modulation added to produce several regularly distributed
local minima.

Our optimizer-directed sampler uses a ``sparsity'' sampling strategy with an ensemble of $16$ Nelder-Mead solvers.

We define our $test$ for validity as:
\begin{equation}\label{eq:test_valid}
    \left(\mathrm{ave}(\Delta_{y}) \leq tol_{ave}\right) \; \land \;
    \left(\max(\Delta_{y}) \leq tol_{max}\right),
\end{equation}
where $\land$ denotes ``and'', $tol_{ave} = 10^{-5}$, $tol_{max} = 10^{-4}$, $\Delta_{x} \ne 0$ is a graphical distance, and $data$ corresponds to all existing
model evaluations (i.e. prior plus newly sampled).
For $train$ we also use Eq.~(\ref{eq:test_valid}) with
$tol_{ave} = 10^{-5}$ and $tol_{max} = 10^{-4}$.
A quality $metric$ for training is given by
$\delta = \sum_{y} \Delta_{y}$ and we define $converged$ as:
\begin{equation}\label{eq:test_converged}
    \Omega(M) \; \vee \;
    \left(\max_{y}(\:\max_{j}(\:\mathrm{ave}(\Delta_{y,j}))) \leq tol_{stop} \right)\; ,
\end{equation}
where $\vee$ denotes ``or'', and $\Omega(M)$ is equal to ``true'' when no new local extrema have been found in the last $M=3$ iterations. 
We use $tol_{stop} = 2 \cdot 10^{-4}$, $\Delta_{y,i}$ is the graphical distance to the $data$ sampled in iteration $i$
(i.e. no prior model evaluations), and $j$ is given by the last $N=3$ iterations $j \in [i-N+1, ..., i]$. 
Using $warm = 1000$, we ensure that at least $1000$ model evaluations are performed per iteration. 
In addition, we track the testing score for a single iteration $i$:
\begin{align}
    score = \mathrm{ave}_{y}(\mathrm{ave}(\Delta_{y,i})),
    \label{eq:test_score}
\end{align}
but do not use it to terminate the calculation. 
To assess stricter tolerances, we repeat the calculation with $tol_{ave} = 10^{-7}$, $tol_{max} = 10^{-6}$, and $tol_{stop} = 2 \cdot 10^{-6}$.
We will refer to these tolerance settings as ``strict'' and the prior tolerance settings as ``loose''.

\begin{figure*}
\includegraphics[width=1\linewidth]{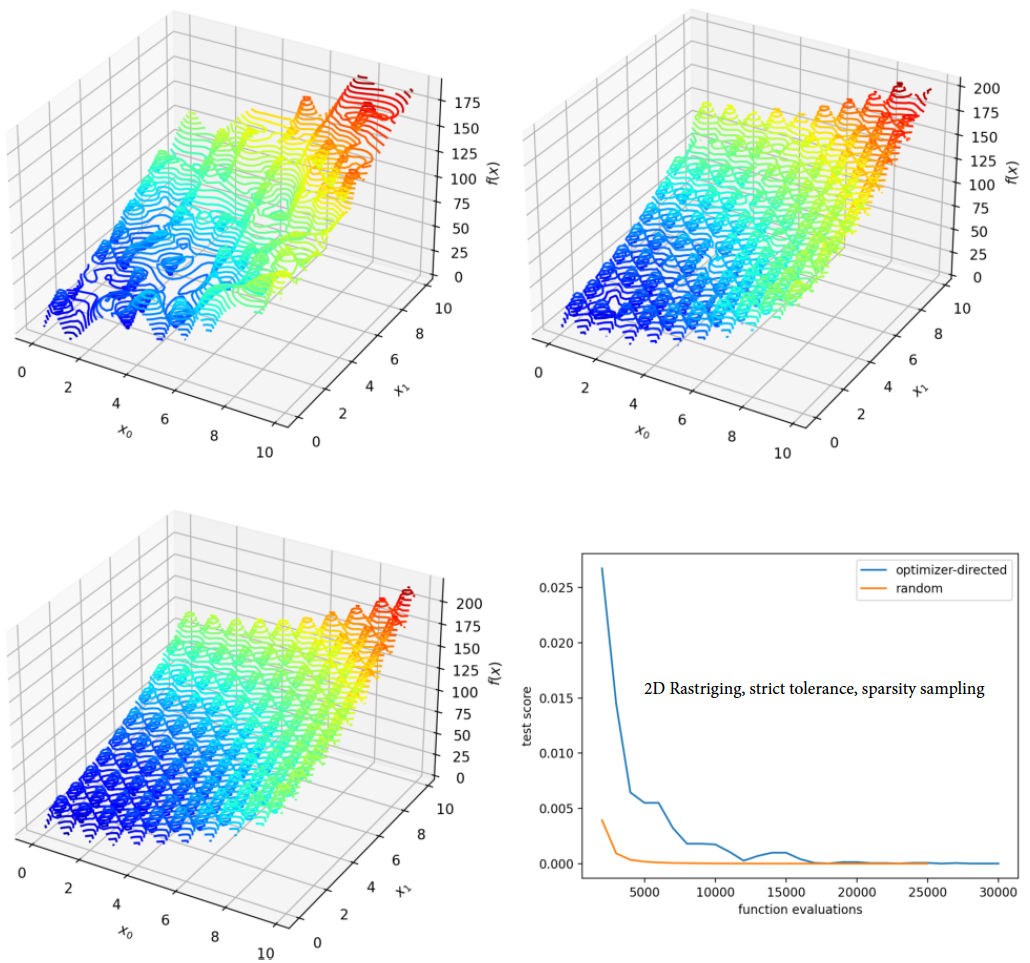}
  \caption{
Candidate surrogates for the 2-dimensional Rastrigin function, learned with a thin-plate RBF estimator using ``sparsity'' sampling, a ``strict'' tolerance, and a test metric for validity based on the average graphical distance between the learned surrogate and sampled data.
Surrogates are plotted with inputs $x = (x_0, x_1)$ and output $z = f(x)$.
Top row: Sampling using ensembles of $16$ optimizers after the initial and tenth iterations. Bottom row: the final iteration and the test score per sample. The final surrogate is visually identical to ground truth and reproduces all local extrema within the desired accuracy. The test score for pure systematic random sampling converges faster than optimizer-directed sampling, as may be expected for a metric based on the average surrogate misfit.
}
  \label{fig:Rastrigin}
\end{figure*}

\begin{figure*}
\includegraphics[width=1\linewidth]{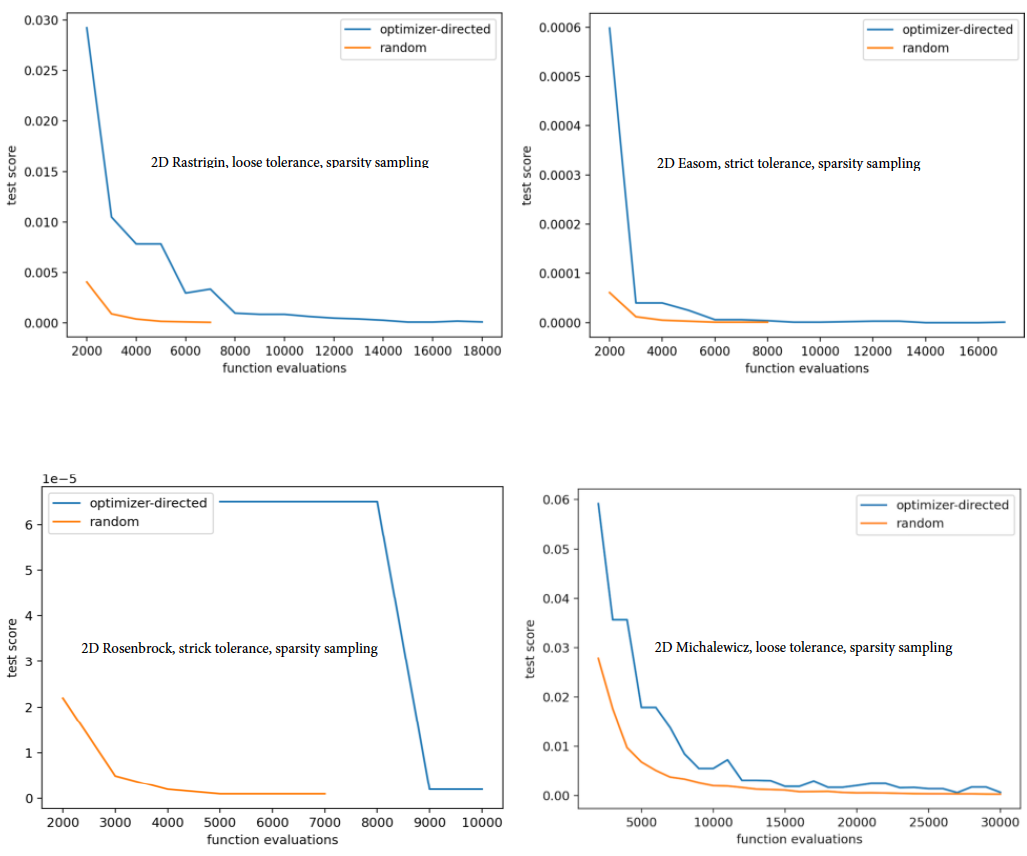}
  \caption{
Convergence of $test$ score versus model evaluations for different benchmark functions learned with a thin-plate RBF estimator using ``sparsity'' sampling, and a test metric for validity based on the average graphical distance between the learned surrogate and sampled data.
Top left: 2-dimensional Rastrigin function with ``loose'' tolerance.
Top right: 2-dimensional Easom function with ``strict'' tolerance.
Bottom left: 2-dimensional Rosenbrock function at strict tolerance.
Bottom right: 2-dimensional Michalewicz function at loose tolerance.
Note the surrogates on the top row reproduce ground truth well regardless of approach, as in Figure \ref{fig:Rastrigin}. Performance of either approach for the 2-D Rosenbrock is as occurs for the 8-D Rosenbrock, in Figure \ref{fig:Rosenbrock8}
Both random and optimizer-directed approaches do equally poorly with the 2-D Michalewicz.
}
  \label{fig:converged}
\end{figure*}

\begin{figure*}
\includegraphics[width=1\linewidth]{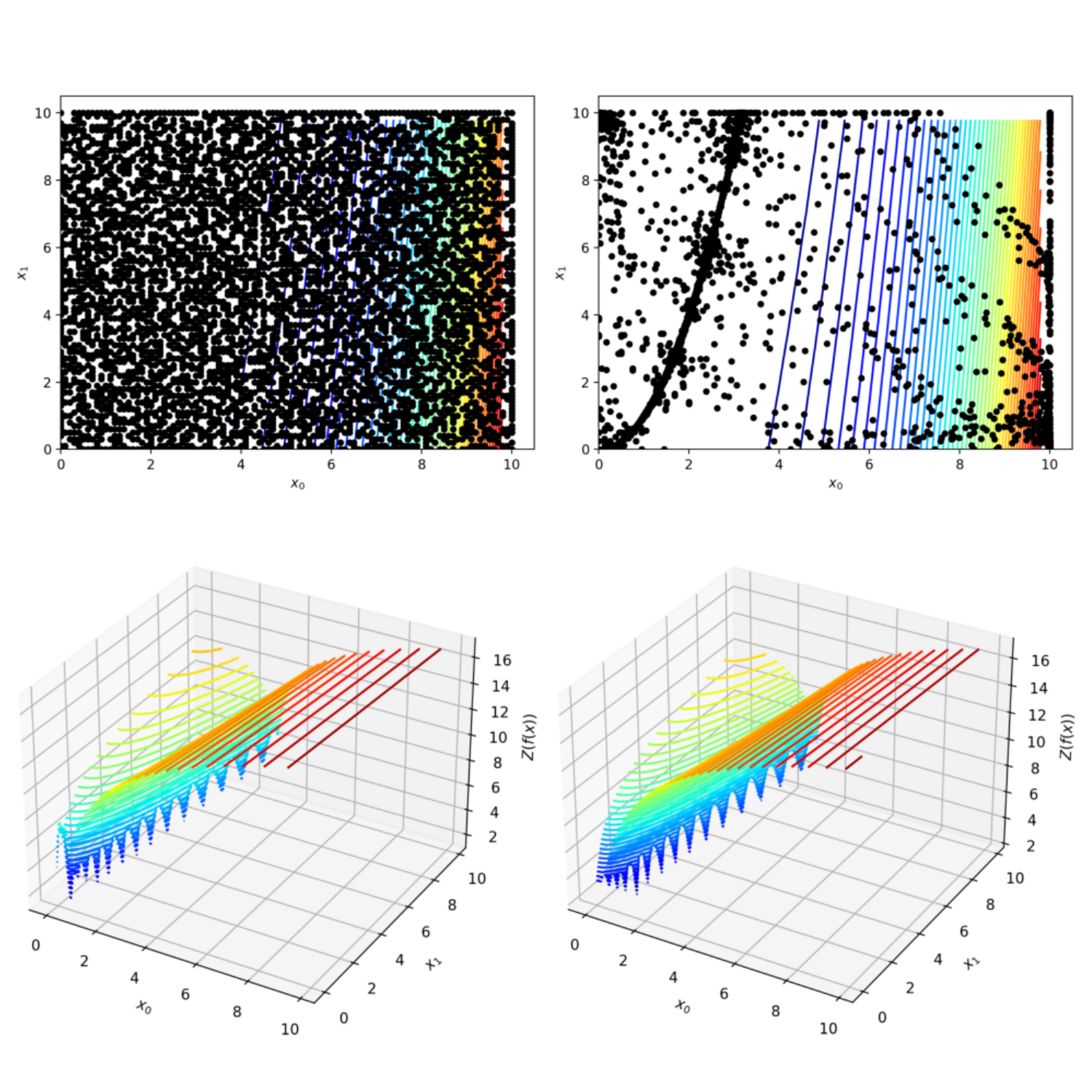}
  \caption{
Candidate surrogates for the 2-dimensional Rosenbrock function, learned with a thin-plate RBF estimator using ``sparsity'' sampling, a ``loose'' tolerance, and a test metric for validity based on the average graphical distance between the learned surrogate and sampled data.
Surrogates are plotted with inputs $x = (x_0, x_1)$ and output $z = Z(f(x))$, where log-scaling $Z = log(4 \cdot f(x) + 1) + 2$ is used to view the region around the global minimum better.
Top row, left: model evaluations sampled with the random sampling strategy.
Top row, right: optimizer-directed sampling. Bottom row, left: log-scaled view of surrogate from random sampling. Bottom row, right: log-scaled view of surrogate from optimizer-directed sampling, which reproduces ground truth.
Note convergence occurs quickly using either strategy, where the $converged$ condition is met with no more than two iterations.
}
  \label{fig:Rosenbrock}
\end{figure*}

\begin{figure*}
\includegraphics[width=1\linewidth]{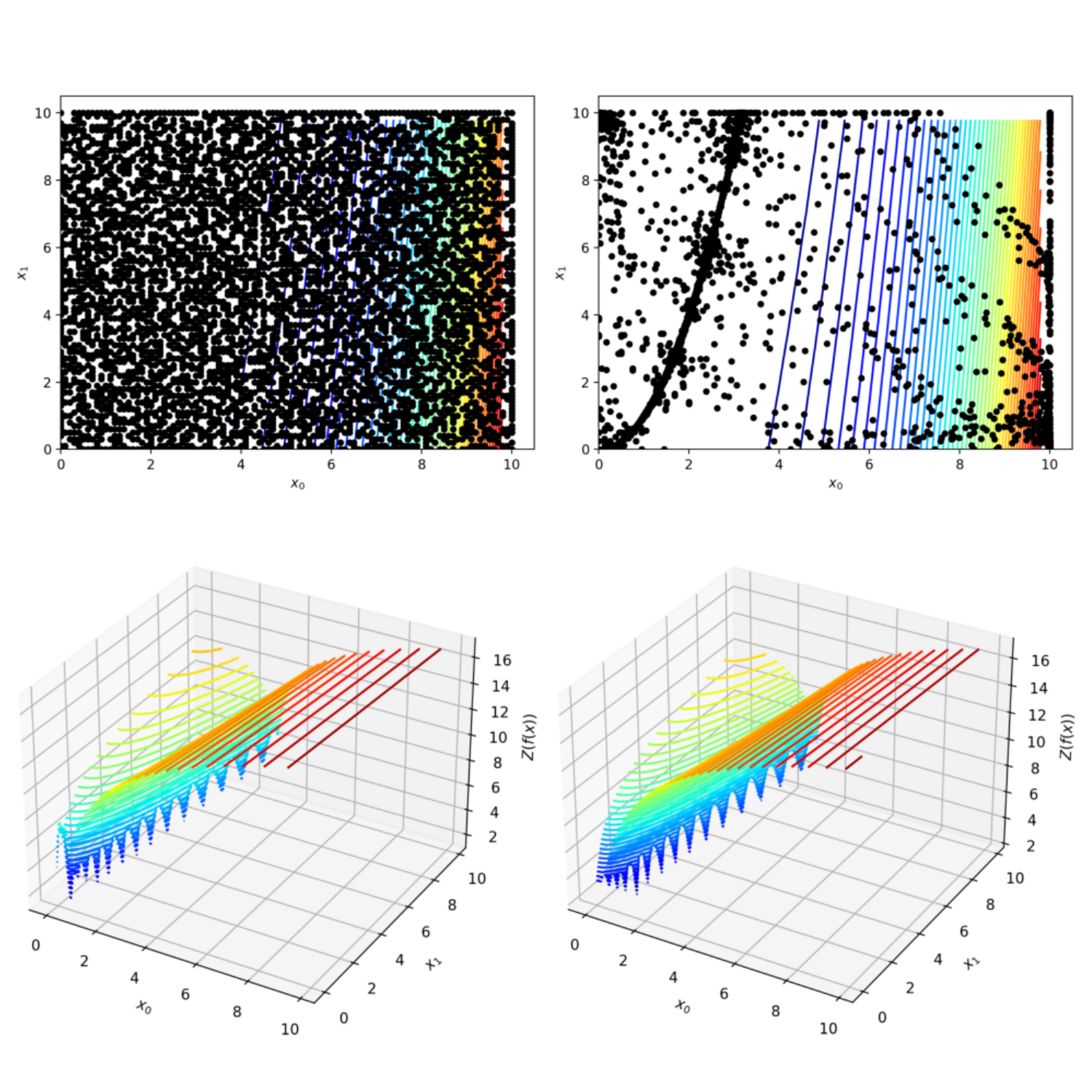}
  \caption{
Candidate surrogates for the 8-dimensional Rosenbrock function, learned with a thin-plate RBF estimator using ``sparsity'' sampling, a ``loose'' tolerance, and a test metric for validity based on the average graphical distance between the learned surrogate and sampled data.
Surrogates are plotted with inputs $x = (x_0, x_1, 1, 1, 1, 1, 1, 1)$ and output $z = f(x)$ or $z = Z(f(x))$, where log-scaling $Z = log(4 \cdot f(x) + 1) + 2$ is used to view the region around the global minimum better. 
Top row, left: test score per sample.
Top row, center: model evaluations sampled with the random sampling strategy.
Top row, right: optimizer-directed sampling. Bottom row, left: surrogates produced with either sampling approach are visually identical to the ground truth. Bottom row, center: log-scaled view of surrogate from random sampling near the global minimum. Bottom row, right: log-scaled view of surrogate from optimizer-directed sampling near the global minimum, identical to the ground truth. While pure systematic random sampling converges faster, optimizer-directed sampling provides a more accurate surrogate near the critical points.
}
  \label{fig:Rosenbrock8}
\end{figure*}

We compare our results with pure systematic random sampling, using an ensemble of $500$ points for strict and loose tolerances. 
As can be seen in Figures \ref{fig:Rastrigin} and \ref{fig:converged} (and in Extended Data Table \ref{table:convergence_tests}),
the test score for pure systematic random sampling converges, yielding an excellent representation of ground truth faster than optimizer-directed sampling (when using the default optimizer configuration and a metric based on the average surrogate misfit) for both strict and loose tolerances.
As the Rastrigin function has shallow local extrema distributed uniformly across the response surface, it may be expected that a systematic random sampling
strategy that efficiently covers input space is more performant than a strategy
that attempts to pinpoint the extrema.

We performed a similar comparison for the $d$-dimensional Rosenbrock function \cite{Rosenbrock:1960}:
\begin{equation}\label{eq:Rosenbrock2}
    f({\bf x}) = \sum_{i=1}^{d-1} 100 (x_{i+1} - x_{i}^{2})^{2} + (1 - x_{i})^{2}
\end{equation}
with $d = \{2, 8\}$,
$d$-dimensional Hartmann's function \cite{Dixson:1978}:
\begin{equation}\label{eq:Hartmann6}
    f({\bf x}) = -\sum^{4}_{i=1} \alpha_i\exp \bigg( -\sum^{d}_{j=1} A_{ij} \big(x_j-P_{ij}\big)^2\bigg)
\end{equation}
with $d = 6$,
$d$-dimensional Michalewicz's function \cite{Michalewicz:1992}:
\begin{equation}\label{eq:Michalewicz2}
    f({\bf x}) = -\sum_{i=1}^{d} \sin(x_{i}) \sin^{20}(i x_{i}^{2} / \pi)
\end{equation}
with $d = 2$,
and
2-dimensional Easom's function \cite{Easom:1990}:
\begin{equation}\label{eq:Easom2}
    f({\bf x}) = -\cos(x_{1}) \cos(x_{2}) \, {\rm e}^{-(x_{1}-\pi)^{2} - (x_{2} - \pi)^{2}}.
\end{equation}
The coefficients $\alpha_i$, $A_{ij}$ and $P_{ij}$ for Hartmann's function can be found in the Supplementary Information.
The 2-D Rosenbrock function is a saddle with an inverted basin, where the global minima occur inside a long, narrow, flat parabolic valley in the inverted basin.
The 8-D Rosenbrock function is the sum of seven coupled 2-D Rosenbrock functions, with a global minimum at $x_{i} = 1$ and a local minima near $x = [-1,1,...,1]$.
The 2-D Michalewicz's function generally evaluates to zero; it has long, narrow channels that create local minima and sharp dips at the intersections of the channels.
The 2-D Easom's function is unimodal and evaluates to zero everywhere except in the region around a singular sharp well.
The 6-D Hartmann's function is similar to Easom's function in that it generally evaluates to zero; however, it is multimodal and composed of several coupled intersecting sharp wells near the origin.

The general form of the saddle in the 2-D Rosenbrock function is captured well
with either systematic random sampling or optimizer-directed sampling, with the former again converging more quickly (as can be seen in Extended Data Table \ref{table:convergence_tests}) for strict and loose tolerances.
However, upon closer examination of the long, narrow channel that
contains the global minimum (see Figure \ref{fig:Rosenbrock}), we find that
optimizer-directed sampling reproduces ground truth much more accurately in the region
surrounding the global minimum. This is to be expected as an optimizer-directed
strategy should provide a higher density of sampling in the neighborhood of the single global minimum. We find similar behavior 
for both strict and loose tolerances and
for the 8-D Rosenbrock function tested with loose tolerances (see Figures
\ref{fig:converged} and \ref{fig:Rosenbrock8}).

The 2-D Easom's and 6-D Hartmann's (Supplementary Information, Benchmark functions) functions provide a similar challenge: they evaluate to zero everywhere, except in the region around one or more sharp wells. Optimizers often terminate shortly after encountering a large plateau on a response surface, so some inefficiency should be expected for optimizer-directed sampling for these functions. While again, optimizer-directed sampling (at the default optimizer configuration and for a metric based on the average surrogate misfit) converges less quickly, the difference is not as lopsided as one might anticipate.

The outlier of the study is the 2-D Michalewicz function. The function is generally flat, with several long, narrow channels that have sharp dips at the intersections of the channels, creating an infinite number of first-order critical points. The infinite number of local minima and maxima causes difficulty for both sampling approaches about efficiency.
We found that either approach visually reproduces ground truth, approximately after $30,000$ model evaluations with loose tolerances.
It is expected that for strict tolerances, much more sampling is required to generate a surrogate for the 2-D Michalewicz function that reproduces the critical points with the quality observed for the other benchmark functions.

In summary, for all cases, systematic random sampling is found to converge faster to a valid surrogate for all future data. However, optimizer-directed sampling better ensures that the behavior at function extrema is reproduced. Note that in all cases, the optimizer used was the Nelder-Mead solver (Supplementary Information,  Default optimizer configuration)
We expect that less strict convergence requirements will reduce the number of evaluations required by the optimizer, potentially at the cost of some accuracy in the vicinity of the critical points. We will explore the impact of optimizer configuration in the following subsection.

\paragraph{\label{sec:TestTraining} {\bf Sampling for Training Validity.}}

Here, we assess the impact on the efficiency of optimizer-directed sampling due to the configuration of the optimizer.
Our optimizer-directed sampler uses a ``lattice'' sampling strategy with
an ensemble of $40$ Nelder Mead solver instances (see Supplementary Information, Sampling model evaluations).
We define our $test$ for validity as:
\begin{equation}\label{eq:test_opt_valid0}
    \sum_{y} \Delta_{y} \leq tol_{sum} \; \land \;
    \max(\Delta_{y}) \leq tol_{max},
\end{equation}
where $tol_{sum} = 1 \cdot 10^{-3}$ and $tol_{max} = 1 \cdot 10^{-6}$. We use a graphical
distance with $\Delta_{x} \ne 0$, and $data$ defined as all existing
model evaluations (i.e., prior plus newly sampled).
We define $train$ as in Eq.~(\ref{eq:test_opt_valid0}), again with
$tol_{sum} = 1 \cdot 10^{-3}$ and $tol_{max} = 1 \cdot 10^{-6}$.
We use a quality $metric$ for training, defined by
$\delta = \sum_{y} \Delta_{y}$, and
define $converged$, in Eq.~(\ref{eq:converged}),
identically to $test$ in Eq.~(\ref{eq:test_opt_valid0}).

 We present (Extended Data Table \ref{table:performance_tests}),
the average time to obtain validity (as defined above) for
surrogates of several standard benchmark functions.
Simulations were performed on the Darwin cluster at Los Alamos National Laboratory on a $36$ core Skylake microarchitecture.
We found for the default optimizer configurations (Supplementary Information, Default optimizer configuration), an ensemble of Nelder-Mead optimizers is more efficient, regardless of the dimensionality of the benchmark function. We surmise this is due to the Nelder-Mead solvers getting stuck in local extrema faster than those using Powell's method. 
As the goal here is for the ensemble of solvers to find as many local extrema as
possible using the minimum number of function evaluations, Nelder-Mead appears to be the better choice.

The following sections will test the ability to quickly produce a valid surrogate reproducing relevant physical behavior in regions where traditional methods have difficulty producing similar results. We will use a larger ensemble of optimizers and test the accuracy at the end of a single iteration of our entire workflow. This does not guarantee the surrogate will be valid against all future data but will give us an idea of how quickly the surrogates can accurately reproduce physical effects near the critical points.

\subsection*{\label{sec:EOS} Equation of State with Phase Transition}

As an illustrative example, we consider building an accurate surrogate for a high-density nuclear-matter equation of state (EOS) that contains a phase transition (PT).  (A second application is described in the Supplement.) Reliable models for nuclear matter exist up to baryon number densities $n_b$ of about twice the nuclear saturation density $n_0 \sim 0.16\: \mathrm{fm^{-3}}$ as well as asymptotically high densities of $n_b > 40 \: n_0$ \cite{Lonardoni20,Eemeli20}.
While at low densities and temperatures $T$, nuclear matter is composed of neutrons and protons, for high values of $n_b$ and $T$ it is expected to undergo a transition to matter composed of deconfined quarks and gluons \cite{Baym2018}. 
There are significant uncertainties regarding the critical temperatures and densities for the onset of the quark phase, giving motivation for heavy-ion experiments \cite{Adam2021,Busza2018, Munzinger2016} and neutron-star research \cite{Raaijmakers2021, Capano2020, Dietrich2020, Riley2019, Miller2019, Dexheimer17, Abbott2017} to study nuclear matter under extreme conditions. 
Corresponding numerical studies need a nuclear EOS, either in analytic or tabulated forms \cite{Typel15, Schneider2019,  Raithel19}. 

To create EOS tables over a large density and temperature range, the most common approach is to select models for the hadronic and quark EOSs and connect them via a 
Maxwell or Gibbs construction to represent the PT \cite{Glendenning97,Hempel09, Fischer11}. 
The Gibbs construction assumes the coexistence of quarks and hadrons in a mixed phase where conservation laws are fulfilled globally.  For the Maxwell construction, only the baryon number is conserved globally.  
Other conservation laws, like electric charge neutrality, are fulfilled separately for quark and hadronic matter.
Neither of these constructions is currently ruled out; in fact, there are other models to connect the pure hadronic and quark phases, for example, via a so-called crossover \cite{McLerran2019, Baym2018}. 
However, the Maxwell construction usually leads to extreme PT characteristics, including a pressure plateau in the mixed phase.
With that, it represents a suitable challenge for a surrogate. 

\begin{figure*}
\includegraphics[width=1\linewidth]{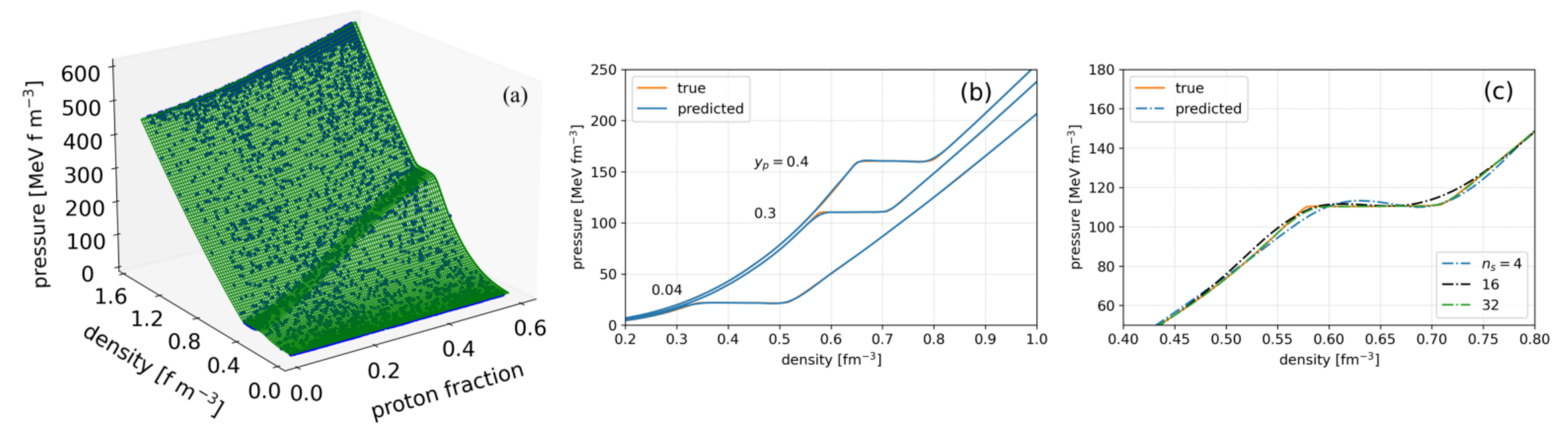}
\caption{
(a) Simulations (dots) and predicted values (surface) of the nuclear matter EOS. The initial search domain $n_b \,\in $ $[0.04 \: \mathrm{fm^{-3}},1.2\: \mathrm{fm}^{-3}]$ and $y_p \in [0.01,0.6]$ was sampled with a lattice sampler. We used $n_s=40$ Nelder-Mead solvers and $test$ validity defined as in Eq.~(\ref{eq:test_opt_valid}) with $tol_{max} = 10^{-6}$, $tol_{sum} = 10^{-3}$, and $train$, $converged$, $data$, and $metric$ as defined in Methods \ref{sec:TestTraining}. A valid surrogate for the EOS that correctly describes the pressure plateau was found after $6846$ model evaluations. (b) Pressure as a function of density for given proton fractions $y_p = 0.04, \: 0.3, \: 0.4$. (c) Pressure as a function of density for $n_b=0.3$, $y_p = 0.3$, and three different sizes of ensembles $(n_s)$. The orange curve shows the results from more expensive simulations, and the blue curve represents prediction using our methodology. As we increase the size of the ensemble of solvers directed to find the critical points, the accuracy of the surrogate obtained from a single learning step using a thin-plate RBF improves. Note the tolerances are absolute, and defined in Eq.~(\ref{eq:test_valid}) with respect to the graphical distance, with $\Delta x \neq 0$.}
\label{fig:EOS1}
 \end{figure*}

We model the PT with the Maxwell construction (see the Supplement). The quark matter EOS is described by the MIT bag model where quarks are non-interacting fermions confined into nucleons at low densities via a confinement pressure, or the so-called bag constant \cite{Chodos74, Fischer11}, taken here to be $B^{1/4} = 170\:\mathrm{MeV}$ (or $B \sim 109 \: \mathrm{MeV/fm^3}$) \cite{Schertler2000}. Quark matter is composed of up, down, and strange quarks, which we take to have masses $m_\mathrm{up}=m_\mathrm{down} = 0$ and $m_\mathrm{strange} = 150$MeV. 
Nuclear EOS tables for astrophysics are often divided into different temperature blocks, with each block containing $y_p$ sub-blocks in which the thermodynamic quantities are provided for a range of densities.  For our demonstration, we assume the pressure to be a function of $n_b$ and $y_p$ only, corresponding to a zero-temperature block. The typical range for $y_p$ in nuclear EOSs is $0.01 \leq y_p \leq 0.6$. To capture the PT, we focus on densities in the range $0.04 \: \mathrm{fm}^{-3} \leq n_b \leq 1.6\: \mathrm{fm}^{-3}$. 

We use lattice sampling with an ensemble of $40$ Nelder-Mead solvers at the default configuration,
and a surrogate learned using a thin-plate RBF.
Here, we defined $test$ validity as in Eq.~(\ref{eq:test_opt_valid}) with $tol_{max} = 10^{-6}$ and $tol_{sum} = 10^{-3}$, and $train$, $converged$, $data$, and 
$metric$ as defined in Section \ref{sec:TestTraining}.
The results are plotted in Fig.~(\ref{fig:EOS1}a), which shows the entire $n_b$-$y_p$ plane, where
the pressure plateau of the PT is visible. 
As seen in previous studies, the critical density for the onset of the mixed phase moves to higher values for increasing $y_p$ \cite{Fischer11}. 
Fig.~(\ref{fig:EOS1}b) gives a more detailed view of the EOS by showing the pressure profiles for fixed $y_p$.  
As can be seen, no systematic errors arise in the predicted pressure values with either proton fraction or density.

The RBFs will reproduce the phase transition accurately if enough of the response surface's inflection points (central to the discontinuity) are sampled. As the optimizers search for all the 
first-order critical points of the surface, the surrogates also begin to reproduce the discontinuities better. 
We illustrated this point in Fig.~(\ref{fig:EOS1}c) for $n_b = 0.3$ and $y_p = 0.3$. As we increase the size of the ensemble of solvers $n_s$
directed to find the critical points, the accuracy of the surrogate obtained from a single learning step using a thin-plate RBF improves.

\section*{Discussion}

We presented an online learning strategy to produce valid surrogates for a chosen quality metric. The approach works well in generating surrogates for existing data and can also be applied about future data. 
We demonstrated an application of online learning where training data is selected with a sampling strategy, and an iterative approach is applied to improve surrogate validity versus the chosen metric. 

We gave evidence that if the critical points of the model's response surface are known, a robust estimator (e.g., thin-plate RBF interpolation or an MLP neural network) should be able to create a surrogate that reproduces the behavior of a more expensive model exactly. We presented an optimizer-directed sampling strategy that effectively samples the critical points of a model's response surface.
We then compared the efficiency of different sampling strategies in learning surrogates that are valid for benchmark functions, even in the presence of newly sampled data. 
Note that if the surrogate was found invalid for newly acquired data, our online approach can be used to improve it iteratively.

We used a sparse sampling approach for selected benchmark functions that produced new draws at the least-populated points in parameter space.
We compared this to an optimizer-directed approach where each initial draw is used as a starting point for an optimizer that runs to termination.
At first blush, it seemed that the traditional sampling strategy outperformed the optimizer-directed one for all benchmark functions (see 
Extended Data Table \ref{table:convergence_tests}).
However, we used a metric based on the error in the surrogate's predicted value versus ground truth, averaged out over the entire response surface.
Hence, it should not be surprising that our methodology produced surrogates that are, on average, highly quality across the entire parameter range.
One might conclude that a traditional sampling strategy, especially one that provides more diffuse sampling than an optimizer-directed strategy, is more efficient at generating valid surrogates when the average misfit across parameter space measures the validity.
We note that the default optimizer configuration was used in testing the efficiency of the sampling strategy, and \emph{tuning} the optimizer may make a substantial difference in the efficiency of the optimizer-directed strategy. 

We also noted that our metric did not guarantee the quality of the surrogate \emph{in the neighborhood of the critical points}.
Returning to our conjecture, finding a response surface's critical points is crucial in guaranteeing the long-term validity of the surrogate as new data is collected.

We found that an optimizer-directed approach is superior at minimizing the model error in the neighborhood of the critical points (see Fig.~\ref{fig:Rosenbrock8}), even when the metric does not call for that explicitly. 
Conversely, a traditional sampling strategy is blind to the response surface and demonstrates a much larger misfit near (at least the first order) critical points. 
Thus, using a metric that judges the quality of the surrogate by the misfit at the critical points should produce high-quality surrogates with an optimizer-directed approach with even greater efficiency.

For a physical system, the critical points of a response surface are usually associated with the occurrence of new phenomena. This provides additional motivation to reduce the misfit near the critical points as much as possible. With that, we applied our methodology to two physics test problems. We showed that we could efficiently learn surrogates for equation-of-state calculations of dense nuclear matter, yielding excellent agreement between the surrogate and model for a wide parameter range and a region that includes a phase transition. 
We also showed that our methodology can produce highly accurate surrogates for radial distribution functions from expensive molecular dynamics
simulations for neutral and charged systems of several dimensions and across an extensive range of thermodynamic conditions (given in the Supplement). While our demonstrations focused on two specific problems, the methodology, and associated code are agnostic to the domain of science and can be utilized for various physics scenarios.

A standard metric that is used to determine the validity of a surrogate is the model error, which is typically a distance such as $\delta = \sum_{y} \Delta_{y}$, or more generally
\begin{equation}\label{eq:error}
\delta = \textit{metric}(\: \hat{y}({\: \bf x}|\xi),\textit{ data})
\end{equation}
with $metric$ being a distance function between the surrogate and all model evaluations, $data$. This definition assesses the quality of the surrogate by measuring its distance from the observed data. Unfortunately, for a small set of observed data that is not representative, any learned surrogate will likely become invalidated by adding new data. 
A potentially more robust quality assessment considers training a surrogate with a statistical metric, such as the \emph{expected} model error. 
It can be defined as taking into account any knowledge about the data-generating distributions (for input and output values) and any uncertainty in the input and output parameters of the model. Complex real-world models are often non-deterministic; thus, an appropriate goal is to either find a surrogate guaranteed to be accurate under uncertainty or a surrogate guaranteed to be robust under uncertainty. With minor adjustments, such as adding a timestamp strategy or invalidating training data, our methodology can be leveraged to build and maintain accurate surrogates for time-dependent models. In future work, we will apply our methods to produce surrogates that are guaranteed to be either accurate or robust under uncertainty and similarly demonstrate the ability to guarantee the accuracy of surrogates for time-dependent models.

\section*{\label{sec:Methods} Methods}

\subsection*{{\bf Surrogate Validity.}}\label{sec:SurrogateValidity}  Our general procedure to create a valid surrogate for an expensive model is shown in Fig.~(\ref{fig:Valid-surrogate}). 
The steps are iterative and include explicit validation and update mechanisms.
To simplify computational complexity, we first link the model to a DB.
Thus, its inputs and output are automatically stored when the model is evaluated. 
The DB of model evaluations is used later to train candidate surrogates.
The corresponding surrogate is retrieved from the surrogate DB and tested for validity during model evaluation.
If no stored surrogate exists, we skip testing and proceed directly to learning a candidate surrogate.
Validity is defined as
\begin{equation}
\textit{test}(\Delta) \text{ is true},
\label{eq:valid}
\end{equation}
where $test$ is a function of the graphical distance, $\Delta$
\begin{equation}\label{eq:graphical}
\begin{split}
\Delta_{y} & = \inf_{{\bf x} \in {\bf \mathcal{X}}} \big|\: \hat{y}({\: \bf x}|\xi) - y\big| + \Delta_{x}, \\
\Delta_{x} & = \big|{\bf x} - {\bf x'}\big| \text{ or } 0,
\end{split}
\end{equation}
with $({\bf x'},y)$ a point in the DB of model evaluations, and ${\bf \mathcal{X}}$ the set of all valid inputs ${\bf x}$ for the surrogate $\hat{y}$ with hyperparameters $\xi$. $\Delta_{x}$ and $\Delta_{y}$ are the pointwise $\Delta$ for $({\bf x'},y)$.
If $\Delta_{x} = 0$, we ignore the distance of the inputs while $\Delta_{y}$ is the minimum vertical distance of point $y$ from the surrogate. 

If Eq.~(\ref{eq:valid}) deems the surrogate valid, the execution stops. 
Otherwise, we update the surrogate by training against the DB of stored model evaluations.
We define validity when training a surrogate similar to Eq.~(\ref{eq:valid}) but with the function $train$ replacing $test$. 

We train the surrogate regarding a quality metric given in Eq.~(\ref{eq:error}). If after training a surrogate has a smaller $\delta$ compared to the current best surrogate, then we store the updated surrogate in the surrogate DB and continue to improve the surrogate until $train$ is satisfied.
If training fails to produce a valid surrogate, we use a sampler to generate model evaluations at new $({\bf x'},y)$ and the process restarts. 

Our general procedure for producing a valid surrogate is extended for
asymptotic validity by adding a validity convergence condition
\begin{equation}\label{eq:converged}
\textit{converged}(\Delta) \text{ is true}
\end{equation}
to be called after the surrogate is deemed $test$ valid,
as in Eq.~(\ref{eq:valid}). Thus, instead of stopping execution when
the surrogate is $test$ valid; the latter merely completes an iteration.
If not $converged$, we trigger a new iteration by sampling new data and continue to iterate until the surrogate validity has $converged$.
This iterative procedure is more likely to generate a surrogate valid for all future data when Eq.~(\ref{eq:converged}) requires some form of convergence behavior for $test$ over several iterations. This is one of the leading innovations of our approach. When the DB of model evaluations is sparsely populated, we expect that any new data will likely trigger a surrogate update.

\subsection*{\label{sec:Interpolation} {\bf Learning Strategy.}}

Our procedure is online, as a sampler can request new model evaluations on the fly, which populate to a DB, and our surrogate is updated by querying the DB and training on the stored model evaluations. The automation of the learning process greatly facilitates online learning.
Our general procedure for automating the production of a valid surrogate is
shown in Fig.~(\ref{fig:Valid-surrogate}) and is extended to asymptotic validity. 
As mentioned earlier, we will use an RBF interpolation to generate our surrogates, where we leverage \emph{mystic} for the automation and quality assurance of surrogate production. The utilization of RBFs arises from their universal capabilities for function approximation and their connection to single hidden-layer feed-forward neural networks (NN) with non-sigmoidal nonlinearities \cite{park1991universal,wu2012using,ROCHA20091573}. Although a multilayer perceptron (MLP) \cite{Rumelhart1985} or another similar NN estimator are also potential choices, we will use RBF interpolation as it is generally more efficient
for online learning \cite{wu2012using}.

Let us assume $y({\bf x})$ is an arbitrary function of vector $\bf{x}$
represented on a subset of $\R^n$, and that the value of $y$ at
input vectors ${\bf x}^j$ $( j=1, \ldots, N)$ are the known $data$
points stored in a DB of model evaluations.
We seek to find a surrogate
$\hat{y}({\bf x})$  with the lowest possible number of the evaluations
\cite{Schaback2007APG,ROCHA20091573,DORVLO2002307} satisfying
Eq.~(\ref{eq:error}). We use Eq.~(\ref{eq:error}),
with $\bf{x'} = \bf{x}$, as opposed to
\begin{eqnarray}
    \hat{y}({\bf x}^j)= y({\bf x}^j)\, \text{for all}\, j=1,\ldots,N
    \label{eq:condition}
\end{eqnarray}
to allow our interpolated surrogates to deviate from the data slightly due to the use of \verb|smooth| and \verb|noise|.
Using a RBF $\phi(r)$, the interpolated surrogate can be written as:
\begin{eqnarray}
        \hat{y}({\bf x}) =\sum^{N}_{j=1} \beta_j \phi(d({\bf x},{\bf x}^j)),
    \label{eq:1}
\end{eqnarray}
where $\beta_j$ are coefficients to be determined,  the thin-plate
($\phi = r^2\ln(r)$) RBF is used, and $d({\bf x},{\bf x}^j)$ is a distance function similar to Eq.~(\ref{eq:error}).
If we choose 
$d({\bf x},{\bf x}^j)=||{\bf x}-{\bf x}^j||$ as the Euclidean distance between an arbitrary vector ${\bf x}$ and ${\bf x}^j$, the values of the coefficient vector $\boldsymbol{\beta}=[\beta_1,\beta_2,\cdots,\beta_N]^T$
are determined
by solving the linear system, ${\bf M} \boldsymbol{\beta} = {\bf Y}$,
where ${\bf M}$ is an $N \times N$ symmetric matrix with elements $M_{ij}=\phi(||{\bf x}^i-{\bf x}^j||)$, and ${\bf Y}=[y({\bf x}^1), y({\bf x}^2),\ldots, y({\bf x}^N)]^T$. To mitigate the potential complications associated with singular matrices in the context of ${\bf M}$ and to introduce a degree of randomness essential for each learned surrogate model, we incorporate a minimal quantity of Gaussian noise into the input data.

\subsection*{{\bf Sampling Strategy.}}\label{sec:Sampling} 

Sampling is an integral part of our online learning workflow. It is used to generate new data points $({\bf x'},y)$ that help inform the learning algorithm whenever training fails to produce a valid surrogate. 
As the goal is an asymptotically valid surrogate, we also use sampling to kick-start a new iteration after Eq.~(\ref{eq:valid}) deems the current iteration's surrogate to be valid (see Fig.~(\ref{fig:Valid-surrogate})).
While our workflow's sampling and learning components are fundamentally independent and can run asynchronously, they are linked through the DB of stored model evaluations. 
The data points generated by the sampler are populated to the DB, while the learning algorithm always uses the data contained in the DB when new training is requested.
If there were no concerns about minimizing the number of model evaluations, we could have samplers run continuously, feeding model evaluations into the DB.
However, as described above, we explicitly include sampling as part of the iterative workflow to minimize the number of model evaluations.

We conjectured that (given the training data) a learned surrogate that, at a minimum, includes all of the critical points of a response surface $y({\bf x})$ is guaranteed to be valid for all future data.
Thus, we postulate that a sampling strategy that uses \emph{optimizer-directed} sampling will be most efficient in discovering the critical points of $y({\bf x})$.
We distinguish \emph{optimizer-directed} sampling from \emph{traditional} sampling. 
Optimizer-directed sampling uses an optimizer to direct the sampling toward a goal.
In contrast, traditional methods, such as simple random sampling, generally ignore the response of $y({\bf x})$.
The utility of simple random sampling is that all the samples will be drawn (with replacement) from a distribution, and thus, all sample points can be chosen simultaneously.
Subsequently, $y({\bf x})$ can be evaluated in parallel for all points drawn in the sampling.
An optimizer-directed approach uses traditional sampling to generate samples for the first draw, then uses each first draw member as a starting point for an optimizer that will direct the sampling of the second and subsequent draws toward a critical point on the response surface.
When an optimizer's termination condition is met, traditional sampling is again used to generate a new starting point for a new optimizer, which proceeds to termination as above.
Thus, while an optimizer-directed strategy may be less efficient in generating new data points, it should be more efficient at finding the critical points of the response surface and thus be the preferred strategy when a surrogate is required to be asymptotically valid. 

\subsection*{\label{ap:TestTraining} Default training validity}

Our optimizer-directed sampler uses a ``lattice'' sampling strategy with
an ensemble of $40$
\verb|NelderMeadSimplexSolver| instances.
We define our $test$ for validity as:
\begin{equation}\label{eq:test_opt_valid}
    \sum_{y} \Delta_{y} \leq tol_{sum} \; \land \;
    \max(\Delta_{y}) \leq tol_{max},
\end{equation}
where $tol_{sum} = 1 \cdot 10^{-3}$ and $tol_{max} = 1 \cdot 10^{-6}$. We use a graphical
distance with $\Delta_{x} \ne 0$, and $data$ defined as all existing
model evaluations (i.e. prior plus newly sampled).
We define $train$ as in Eq.~(\ref{eq:test_opt_valid}), again with
$tol_{sum} = 1 \cdot 10^{-3}$ and $tol_{max} = 1 \cdot 10^{-6}$.
We use a quality $metric$ for training, defined by
$\delta = \sum_{y} \Delta_{y}$, and
define $converged$
identically to $test$ in Eq.~(\ref{eq:test_opt_valid}).

\section*{Data availability}
The dataset to create the figure plots in the main text and Supplementary Information can be found on Zenodo  \cite{diaw2024efficient}.

\section*{Code availability}
The code, as well as the sampled data and learned surrogates, is available on Code Ocean \cite{Diaw2024software}.

\section*{Acknowledgments}
Research supported by Los Alamos National Laboratory under the Laboratory Directed Research and Development program (project numbers 20190005DR, 20200410DI, and 20210116DR), by the Department of Energy Advanced Simulation and Computing under the Beyond Moore's Law Program, and by the Uncertainty Quantification Foundation under the Statistical Learning program. Triad National Security, LLC operates the Los Alamos National Laboratory for the National Nuclear Security Administration of the U.S. Department of Energy (Contract No. 89233218CNA000001). M.S.M. acknowledges support from the National Science Foundation through award PHY-2108505. The authors thank Jeff Haack for insightful feedback on the manuscript. This document is LA-UR-20-24947.

\section*{Author contributions statement}
A.D., M.M., and M.S.M. conceived the project. M.M. developed the software.
A.D., I.S., and M.M. performed simulations and prepared figures. All authors were responsible for the formal analysis.

\section*{Competing financial interests:}  The authors declared no competing financial interests. 
\section*{Additional information:} 
\section*{Extended data} 

This section contains the convergence and performance test cases for the benchmark functions.

\setcounter{table}{0} 

\renewcommand{\tablename}{Extended Table}

\begin{table*}[!t]
\centering
\def\arraystretch{1.5}

\begin{subtable}{.45\textwidth}
\centering
\resizebox{\textwidth}{!}{
\begin{tabular}{|c|c||c|c||c|c|}
 \hline
 \multirow{2}{*}{Function} &
 \multirow{2}{*}{ndim} &
 \multicolumn{2}{c||}{Random} & 
 \multicolumn{2}{c|}{Optimizer-directed} \\ 
 \cline{3-4}
 \cline{5-6}
     &  & loose & strict & loose & strict \\
 \hline
 \hline
 Easom       & 2 & 2000 & 8000 & 2939 & 17967 \\ 
 \hline
 Rosenbrock  & 2 & 2000 & 7000 & 7317 & 12111 \\ 
 \hline
 Rastrigin   & 2 & 7000 & 25000 & 18579 & 32308 \\ 
 \hline
 Michalewicz & 2 & 30000 & -- & 30696 & -- \\ 
 \hline
 Hartmann    & 6 & 11000 & -- & 26411 & -- \\
 \hline
 Rosenbrock  & 8 & 15000 & -- & 31487 & -- \\ 
\hline
\end{tabular} 
}
\caption{} 
\label{table:convergence_tests}
\end{subtable}%
\begin{subtable}{.45\textwidth}
\centering
\resizebox{\textwidth}{!}{
\begin{tabular}{|c|c|c||c|c|}
 \hline
 \multirow{2}{*}{Function} &
 \multirow{2}{*}{ndim} &
  \multirow{2}{*}{bounds} &
 \multicolumn{2}{c|}{Function evaluations} \\ 
\cline{4-5}
     &  & & Powell & Nelder-Mead \\
 \hline
 \hline
 Ackley & 2  &  [-1,1] &  3746 &  1631  \\ 
  \hline
 Branins & 2  &  [-10,20] & 2767  &  1007  \\ 
 \hline
 Rosenbrock        & 3 &  [-3, 3] &  4796 & 1733   \\ 
 \hline
  Michalewicz        & 5 &  [0, 3]&  12745 &   1116\\ 
 \hline
Hartmann    &  6 & [-1, 1] & 11896 &  1393 \\ 
  \hline
Rosenbrock &  8 & [-6, 6] & 18430 & 1185   \\ 
\hline
\end{tabular} 
}
\caption{} 
\label{table:performance_tests}
\end{subtable}
\caption{Combined table of convergence and performance tests for benchmark functions.
(a) show the number of evaluations required for several benchmark functions to reach $tol_{stop}$ for both ``loose'' and ``strict'' tolerances,
using a \texttt{SparsitySampler} with bounds ${\bf x} \in [0,10]$. In all cases, systematic random sampling converges more quickly to a valid surrogate for all future data. Using optimizer-directed sampling, however, ensures that the function extrema are known.
The optimizer was a Nelder Mead simplex solver at the default configuration. Less strict convergence requirements will reduce the number of evaluations the optimizer requires, potentially at the cost of some accuracy near the extrema.
Note: For the Michalewicz, Hartmann, and Rosenbrock functions in Table (a), the 'strict' column values are omitted as we did not conduct these evaluations due to the anticipated high computational expense of these cases.
 (b) compares time to validity for several benchmark functions,
using a lattice sampler
with an ensemble of $4$ optimizers
with the default configuration.
We present the number of function evaluations
required to find a valid surrogate, where $converged$
and $test$ are defined as in Eq \eqref{eq:test_opt_valid},
with $tol_{sum} = 10^{-3}$ and $tol_{max} = 10^{-6}$. The sampler is configured to run until all of the optimizers in the ensemble have terminated.
}
\label{table:combined}
\end{table*}

\section*{Supplementary information}  The online version contains supplementary material. 

\bibliographystyle{naturemag-doi} 
\bibliography{references}
\end{document}